\documentclass[11pt]{article}
\usepackage{latexsym}
\usepackage{color}
\usepackage{amssymb, graphicx, amsmath}
\usepackage{amsthm}
\usepackage{wrapfig}
\usepackage{subfigure}
\usepackage{graphicx}
\usepackage{url}
\usepackage{CJK,graphicx}
\usepackage{algorithm}
\usepackage{multirow}
\usepackage{algpseudocode}
\usepackage{subfigure}
\usepackage{float}
\usepackage{appendix}

\usepackage{algorithm}

\usepackage{amssymb}% http://ctan.org/pkg/amssymb
\usepackage{pifont}% http://ctan.org/pkg/pifont

\usepackage{amsfonts,amsmath,amsthm,amsfonts,amssymb}
\usepackage{graphics,epsfig,graphicx,subfigure}
\usepackage[table]{xcolor}
\usepackage{helvet,epsfig,graphicx,subfigure}
\usepackage[table]{xcolor}

\usepackage{textcomp}

\usepackage{tcolorbox}
\usepackage[normalem]{ulem}

\theoremstyle{definition}

\theoremstyle{remark}

\numberwithin{equation}{section}

\usepackage{algorithm}
\usepackage{tikz}
\usepackage{algpseudocode}
\usepackage{amsmath}  
\usepackage{booktabs,array,xcolor,colortbl,multirow,epsfig,graphicx,subfigure,amsmath,amsthm,amsfonts,amssymb,bm}

\begin{document}
%\begin{CJK*}{GBK}{song}

\title{\Large Heterogeneous Graph-based Knowledge Transfer for Generalized\\[4pt] Zero-shot Learning}

\author{Junjie~Wang\thanks{School of Computer Science and Technology, East China Normal University, Shanghai 200092, China. (E-mail: jasonwang.ecnu@gmail.com)}, Xiangfeng~Wang\thanks{School of Computer Science and Technology, East China Normal University, Shanghai 200062, China. (E-mail: xfwang@cs.ecnu.edu.cn)}, Bo~Jin\thanks{School of Computer Science and Technology, East China Normal University, Shanghai 200062, China. (E-mail: bjin@cs.ecnu.edu.cn)}, Junchi~Yan\thanks{Department of Computer Science and Engineering, Artificial Intelligence Institute, Shanghai Jiao Tong University, Shanghai 200240, China. (E-mail: yanjunchi@sjtu.edu.cn)}, Wenjie~Zhang\thanks{Tecent, Shenzhen 518000, China. (E-mail: izhangwenjie@gmail.com)}, and~Hongyuan~Zha\thanks{School of Computer Science and Technology, East China Normal University, Shanghai 200062, China. (E-mail: zha@cs.ecnu.edu.cn)}
%~\IEEEmembership{Member,~IEEE,}
% <-this % stops a space
%\thanks{J. Wang, X. Wang, B. Jin, and H. Zha are with the Department of Computer Science and Technology, East China Normal University, Shanghai 200092, China. J. Yan is with the Department of Computer Science and Engineering, Shanghai Jiaotong University, Shanghai 200240, China. W. Zhang is with Tecent, Shenzhen 518000, China.}
}

\date{\today}

\maketitle

\begin{abstract}
Generalized zero-shot learning (GZSL) tackles the problem of learning to classify instances involving both seen classes and unseen ones. The key issue is how to effectively transfer the model learned from seen classes to unseen classes. Existing works in GZSL usually assume that some prior information about unseen classes are available. However, such an assumption is unrealistic when new unseen classes appear dynamically. To this end, we propose a novel heterogeneous graph-based knowledge transfer method (HGKT) for GZSL, agnostic to unseen classes and instances, by leveraging graph neural network. Specifically, a structured heterogeneous graph is constructed with high-level representative nodes for seen classes, which are chosen through Wasserstein barycenter in order to simultaneously capture inter-class and intra-class relationship.
The aggregation and embedding functions can be learned through graph neural network, which can be used to compute the embeddings of unseen classes by transferring the knowledge from their neighbors.
%More importantly, in contrast to traditional semantic space embedding techniques, the proposed approach can also effectively alleviate the hubness problem ([[[ what is this?]]]).
Extensive experiments on public benchmark datasets show that our method achieves state-of-the-art results.
\end{abstract}

\section{Introduction}
%Human can identify a new object via very few samples or based on highly abstract descriptions. For example, if parents tell their child that a zebra has the tiger-like pattern, the panda-like color and the horse-like shape (Fig. \ref{fig:intro}), the child can recognize the zebra even though he has never seen it. In contrast, to learn a well-performing classifier, most existing models usually require plenty of labeled instances. However, due to huge category number and labeling difficulty, it can be tedious or even impossible to label instances for each class. As a result, Zero-Shot Learning (ZSL) that can handle samples falling into unseen categories has recently received intensive attention \cite{palatucci2009zero}.

\begin{wrapfigure}{r}{8.5cm}
  \centering
  \includegraphics[width=0.85\linewidth]{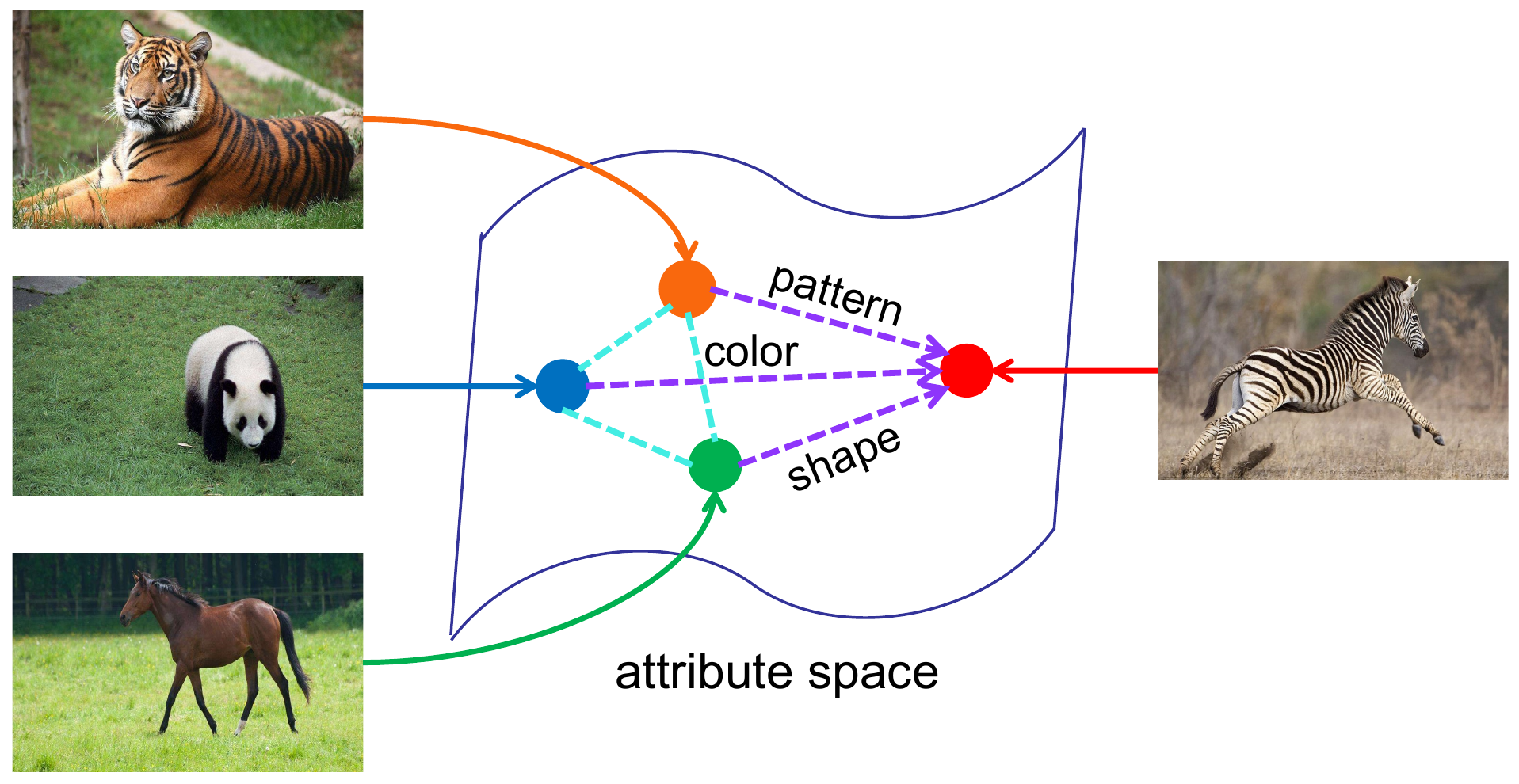}
  \vspace{-10pt}
  \caption{The unseen class in training set (zebra) and seen classes (tiger, panda and horse) can be described by common semantic information (attribute: pattern, color and shape).}
  \label{fig:intro}
\end{wrapfigure}

Zero-Shot Learning (ZSL) that can handle samples falling into unseen categories has recently received intensive attention~\cite{palatucci2009zero,xian2018zero,wang2019survey}. The key challenging problem of ZSL is how one can correctly classify those instances from unseen classes which are absent during training? In general, the instances can be described by some common high-level semantic information, for instance the pattern, color and shape in Fig. \ref{fig:intro}. One viable approach to deal with the unseen classes is to transfer knowledge learned from seen classes exploring those common semantic information mentioned above. Following this approach, the success of ZSL in general depends on the following two factors: (1) how to capture relationship between all seen and unseen classes? (2) how to transfer knowledge based on this relationship? 

Effective modeling of the relationship is of vital importance to knowledge transfer, which has been intensively studied in literature. One typical way is to use the high-level semantic information to bridge seen classes and unseen ones. The attribute vector for each class is usually called {\bf{class prototype}}, It is one of the most widely used semantic information to construct the relationship. For example, in prototypical network \cite{snell2017prototypical}, the visual features are  directly mapped into attribute space, and the nearest class prototype is selected as its label. However, these approaches only introduce the same semantic space to represent both seen and unseen classes without capturing relationship explicitly. Another way is to construct relationship explicitly by using knowledge graph (KG) \cite{kampffmeyer2019rethinking}. Nevertheless, unseen class information is required to construct the graph before training, hence the model need to be retrained every time a new unseen class appears. 

%In this paper, we look into a more realistic scenario, generalized zero-shot learning (GZSL). Compared to ZSL only search in unseen class at test time, GZSL has to decide between both seen and unseen class. We proposed a heterogeneous graph embedding model (HGKT) that is agnostic to both unseen images and unseen semantic vectors during training. Specifically, we focus on image classification task and construct a fancy graph can capture intra-class and inter-class relationship simultaneously. In particular, each node represent for one instance in our graph. We construct a complete-graph for each class to capture intra-class relationship, which means we hope that instances in same class have similar representation in embedding space.

In this paper, we propose a novel and explainable way to utilize the relationship between seen and unseen classes for ZSL. In particular, we address a more realistic scenario, which involves the inference of both seen and unseen classes at testing phase --- generalized zero-shot learning (GZSL). Furthermore, our method is agnostic to both unseen images and unseen semantic vectors during training phrase, thus enjoying more flexibility in practice compared with many previous methods~\cite{wang2018zero,kampffmeyer2019rethinking}.

\begin{figure*}[tb!]
	\centering
	\includegraphics[width=0.925\textwidth]{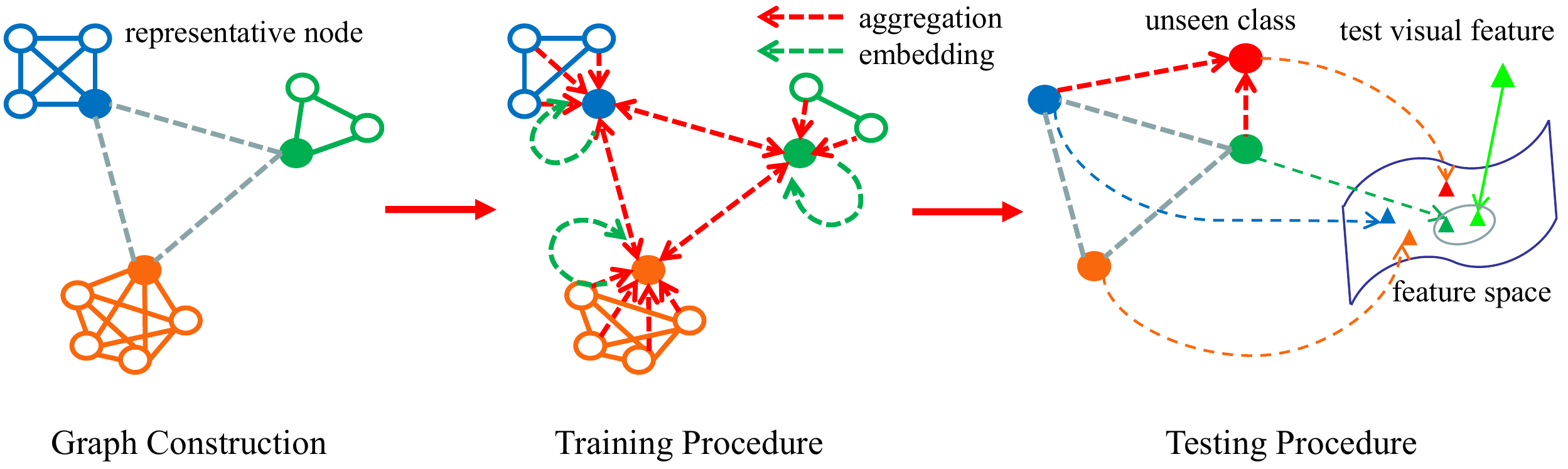}
	\vspace{-10pt}
	\caption{Pipeline of our approach. Left: each class corresponds to a complete graph and all complete subgraphs are connected based on their representative nodes (solid circle). Middle: the embedding vector of representative nodes are produced by both aggregation and embedding functions which is agnostic to to unseen class. Right: connect the new unseen class with $k$-nearest seen classes in visual feature space; select the nearest class by visual feature distance as the prediction for each test sample.}
	\label{fig:framework}
\end{figure*}

To this end, we construct a graph with heterogeneous structure to capture intra-class and inter-class relationship simultaneously, each node representing one instance. In order to capture intra-class relationship, we construct a complete graph for all the instances in each class, which can force instances to be as close as possible in embedding space. More importantly, more similar classes should be closer in embedding space, thus we also need to construct inter-class relationship for transferring knowledge between classes. However, there are many instances in each class, thus it is essential to select representative nodes to represent the whole class. Inspired by Generative Adversarial Networks (GAN)~\cite{goodfellow2014generative}, the instances belonging to the same class are expected to be subject to the same distribution.
As a result, the normalized feature vector of each instance can be considered as a discrete vector sampled from this common distribution.
The Wasserstein metric~\cite{villani2008optimal} is a natural measure of probability distribution. Thus the representative node of each class is selected as the one nearest to the barycenter, while the distance is defined by the Wasserstein metric. %\textcolor{blue}{In our experiments, we will present some quantitative results and visualization to show advantages of Wasserstein barycenter.} 
All the seen classes are connected by linking these representative nodes following kNN scheme. Based on the intra-class and inter-class connections, we obtain the heterogeneous graph with two node types on the seen classes as shown on the left of Fig. \ref{fig:framework}.

Another important issue is to transfer knowledge based on the well-constructed heterogeneous graph. We focus on transferring intra-class and inter-class knowledge simultaneously. Two problems need to be solved, i.e., which embedding space to choose and how to technically transfer information on the graph. Most existing models choose either the semantic space or an intermediate space as embedding space, which usually lead to the so-called hubness problem \cite{shigeto2015ridge}, i.e., there always exist some points which are 'universal' neighbors or hubs for most of all other points. To alleviate this problem, we use the output visual feature space as the embedding space. Moreover, the popular graph neural network (GNN) is utilized to transfer knowledge rely on the heterogeneous graph, which maintains the topology in embedding space. Especially, the aggregation and embedding functions are calculated by using seen training data, further we can obtain the embedding expression of new unseen classes by aggregating their neighbors.
The general algorithm framework can be found in Fig. \ref{fig:framework}.

To summarize, we propose an effective, practical and explainable GZSL algorithm with key contributions as:
1) We capture inter-class and intra-class relationship jointly by constructing a heterogeneous structured graph;
%devise an effective, practical and explainable way to capture inter-class and intra-class relationship simultaneously;
2) Instead of averaging instances directly, we utilize Wasserstein metric to extract more  representative node of each class;
%3) {\color{blue}{The connection between seen and unseen classes are established using kNN, knowledge is transferred from seen to unseen classes by using well-learned aggregate and embedding functions.}}[[[]how transfer is done]]]
3) Our approach is the novel inductive GNN-based GZSL method that is agnostic to unseen information during training.
Knowledge is transferred from seen classes to new unseen classes based on the learned aggregation and embedding functions.

\section{Related Work}
There has recently been intensive works on generalized zero-shot learning. Most works rely on side information such as attributes to establish connection \cite{fu2018recent}.
Based on these well-defined attributes, traditional algorithms establish the bridge between visual features and class prototype. These approaches can be viewed from the perspective of embedding space and embedding model.

\subsubsection{Embedding Space.}
One of the most intuitive embedding space is the semantic space, i.e., mapping the feature space directly into attributes space. The work \cite{lampert2014attribute} proposes Direct Attribute Prediction (DAP) and Indirect Attribute Prediction (IAP) respectively, which both learn the mapping from visual feature space to attribute space via support vector machine (SVM).% Moreover, in \cite{akata2016label}, the proposed algorithm establishes the mapping from features to attributes directly under the ranking loss.

Another popular way is to map both features and attributes into an intermediate space. The authors in \cite{zhang2015zero} consider each source or target data sample as a mixture of seen class proportions, and moreover it's natural to map features and attributes into the simplex of seen classes by measuring similarity. %Similarly, \cite{liu2018generalized} proposes a Deep Calibration Network (DCN) algorithm to learn a jointly embedding space by combining visual image features and semantic representations of each class prototype. 
Many algorithms are devised following this direction, e.g., \cite{romera2015embarrassingly,yang2018dissimilarity}. 

Another line of research is to oppositely embed attributes into the visual feature space. The work \cite{shigeto2015ridge} demonstrates the existence of some `universal' neighbors or hubs, which indicates that kNN based methods would suffer from the hubness problem. %While \cite{dinu2014improving} utilized the global feature instance distribution of unseen data to ease the hubness problem, it is thus transductive. In contrast, 
The work \cite{zhang2017learning} shows how to embed attributes into the visual feature space inductively, which can mitigate both hubness problem.

\subsubsection{Embedding Model.}
From the perspective of embedding modeling, we can divide the algorithms into two categories, i.e., non-deep embedding and deep embedding.
%DAP and IAP algorithms are typical non-deep embedding algorithms, in which SVM is employed to learn the embedding.
In \cite{qi2017joint}, the canonical correlation analysis (CCA) \cite{hardoon2004canonical} is  introduced to train the embedding which achieves significant improvement.    

Meanwhile, many models \cite{frome2013devise,annadani2018preserving,zhang2017learning} construct deep neural networks to map visual feature space to semantic space.
%{\color{red}{Inspired by meta-learning, \cite{hu2018correction} establishes a correlation network for GZSL.}}
They may not directly introduce CNN to learn the embedding, but extract visual feature from original image through CNN. These well-trained deep features usually achieve significant performance. %Recently, \cite{zhu2018generalized} proposed to use graph probability model which is agnostic to both unseen images and unseen semantic vectors during training procedure. However, because the model need to decompose images into visual parts, it can only adapt to some specific datasets.
%{\color{red}{Moreover, \cite{Long2017Zero} use original images instead of extracted features to train Generative Adversarial Networks (GAN).}}

In particular, graph neural network \cite{Franco2009The} has recently received great attentions for their flexibility on dealing with graph structures. Inspired by graph convolutional network, the work 
\cite{wang2018zero} introduces GNN to transfer visual classifier of seen classes to unseen classes based on a given knowledge graph. However, these methods need to know the correlation between seen and unseen classes which is hard to obtain before training.

In this paper, we utilize the inter-class and intra-class relationship by constructing a heterogeneous graph. Specifically, for each node we use the neighbor information to obtain the embedding result through graph neural network, and then embed unseen class based on the trained embedding model. Compared with previous work in \cite{wang2018zero}, the correlation between all classes need not to be known in advance in our approach which improves its practical utility. The proposed method is inductive. In addition, we simultaneously use the visual features of images as the embedding space to alleviate the hubness problem.

\section{The Proposed Approach}
The goal is to transfer information about the seen classes to unseen classes, and the main obstacle is how to extract and deliver information properly. Our main work is to learn aggregation and embedding functions exactly based on GNN. We also introduce some basic information of Wasserstein barycenter and GNN. The proposed method to deal with zero-shot learning problems by utilizing GNN will be detailed inductively as follows.

\subsection{Preliminaries}
\subsubsection{Notations.}
Given a training set $D_{tr} = \left\{ (x_i,y_i,p_i) \right\}_{i=1}^N$, where $x_i \in {\mathbb{R}}^n$ is the visual features of $i$-th image (here we assume it is extracted by a pretrained network e.g. ResNet \cite{he2016deep} which is also a common practice in ZSL literature \cite{xian2018zero}), $y_i$ is the label of $i$-th image within seen class $\mathcal{Y}^{tr}$, i.e., $y_i \in \{1, \dots, L\}$. For each class corresponding to an attribute vector, $p_i \in \mathbb{R}^d$ is the class prototype of the $i$-th image. The goal is to learn a function using $D_{tr}$ and then predict the label for test visual feature $x$ within the whole class dictionary $\mathcal{Y}^t$ including both seen classes $\mathcal{Y}^{tr}$ and unseen classes $\mathcal{Y}^{ts}$, where $\mathcal{Y}^{tr} \cap \mathcal{Y}^{ts} = \emptyset$. Moreover, $\sum_k$ refers to the probability simplex with $k$ bins, and for two matrices of the same size $A$ and $B$, $\left\langle A, B \right\rangle := trace(A^{\rm T} B)$ is the Frobenius dot-product. $I$ refers to the identity matrix.

\subsubsection{Wasserstein Metric.}
Wasserstein metric can be an effective way to measure difference between probability distribution. Given two probabilities ${\mathbf{a}}$ and ${\mathbf{b}}$ and the non-negative cost matrix ${\mathbf{C}}$, we can define the 1-Wasserstein distance between ${\mathbf{a}}$ and ${\mathbf{b}}$ as follows:
\begin{equation}
{{\mathcal{W}}_{\mathbf{C}}}({\mathbf{a}},{\mathbf{b}})\mathop := \mathop {\min }\limits_{ {\mathbf{P}} \in {\rm{U}} ({\mathbf{a}},{\mathbf{b}})}  \left\langle {\mathbf{P}}, {\mathbf{C}} \right\rangle,
\end{equation}
where ${\mathbf{P}} \in \mathbb{R}^{n \times m}$ is transport matrix, and ${\mathbf{P}}_{ij}$ indicates how much mass has moved from position $i$ to position $j$. ${\rm U}({\mathbf{a}},{\mathbf{b}})$ is the feasible region, which can be defined by:
\begin{equation}
    {\rm U}( {\mathbf{a}}, {\mathbf{b}}) := \left\{ {\mathbf{P}} \in {\mathbb{R}}_+^{n \times m}: {\mathbf{P}} {\mathbf{1}}_m = {\mathbf{a}},\ {\mathbf{P}}^{\rm T} {\mathbf{1}}_n = {\mathbf{b}} \right\}.
\end{equation}

A basic problem in machine learning is to compute the `mean' or `barycenter' of several data points. %In Euclidean space $\mathbb{R}^d$, we can define such a weighted mean of points $\left\{ x_s \right\}_{s=1}^{S}$ by solving variational problem:
%\begin{equation}
%\min_{x \in \mathbb{R}^d} \sum_{s=1}^{S} \lambda_s \left\|x-x_s\right\|_2^2, \label{barycenter}
%\end{equation}
%for a given family of weights $\left( \lambda_1,\cdots,\lambda_S\right) \in \sum_S$. 
Based on the above Wasserstein distance, it is natural to introduce the Wasserstein barycenter \cite{agueh2011barycenters}. Given parameter $\lambda$ and input histogram $\{{\mathbf{b}}_s\}_{s=1}^S$, where ${\mathbf{b}}_s \in \sum_{n}$, a Wasserstein barycenter is computed by minimizing
\begin{equation}
{\mathcal{WB}}(\lambda, \{{\mathbf{b}}_s\})=\arg\min_{{\mathbf{a}} \in \sum_n} \sum_{s=1}^{S} \lambda_s {\mathcal{W}}_{\mathbf{C}} ({\mathbf{a}}, {\mathbf{b}}_s), \label{barycenter}
\end{equation}
based on the specified cost matrix $\mathbf{C} \in \mathbb{R}^{n \times n}$.

%However, due to memory and computation complexity, previous solutions such as interior point method and network simplex method become impractical. people usually use Sinkhorn algorithm \cite{cuturi2013sinkhorn} to get approximate solution.
%\begin{equation}
%{{\mathcal{L}}_{\mathbf{C}}^{\epsilon}} ({\mathbf{a}},{\mathbf{b}}) \mathop := \mathop {\min }\limits_{{\mathbf{P}} \in {\rm U} %({\mathbf{a}},{\mathbf{b}} )} \left\langle {\mathbf{P}},{\mathbf{C}} \right\rangle - \epsilon {\mathcal{H}}({\mathbf{P}}), \label{sinkhorn}
%\end{equation}where $\epsilon$ denotes the penalty parameter and ${\cal{H}}$ denotes the entropy regularization, i.e.,
%\begin{equation}
%    {\mathcal{H}} ({\mathbf{P}}) = - \sum_{i,j} {\mathbf{P}}_{i,j} \left(\log \left( {\mathbf{P}}_{i,j} \right) - 1 \right).  
%\end{equation}
%Since the objective (\ref{sinkhorn}) is a $\epsilon$-strongly convex function, thus a unique optimal solution $P_{\epsilon}$ can be guaranteed.
%Moreover, \cite{peyre2019computational} proved that when $\epsilon \rightarrow 0$, then $P_{\epsilon} \rightarrow P$.
%With the help of Sinkhorn technique, the approximation formulation of (\ref{barycenter}) can be expressed as
%\begin{equation}
%\min_{{\mathbf{a}} \in \sum_n} \sum_{s=1}^{S} \lambda_s {\mathcal{L}}_{C}^{\epsilon} ({\mathbf{a}}, {\mathbf{b}}_s).
%\end{equation}

\subsubsection{Graph Neural Network.}
%GNN is first formally introduced in \cite{Franco2009The}. 
We modify GraphSAGE \cite{hamilton2017inductive} to solve our problem. The main idea of GraphSAGE is to learn an aggregation function to collect the information of the node's local neighbors, and learns the embedding function based on the neighbor information. By denoting the learning procedure as function $g$, the main goal is to minimize the following empirical risk:
\begin{equation}
    \min\limits_{g \in \mathcal{G}} \frac{1}{M}\sum_{m=1}\limits^{M} \ell \left( y_m, g \left( x_m \right) \right),
\end{equation}where $\ell$ usually denotes the $softmax$ loss function, which measures the loss incurred from prediction $g(x)$ to the true label $y$.
%For the reason that $g$ and $h$ are both differentiable, thus traditional back propagation technique can be used to train the model.
During testing time, the labels of other $N$-$M$ instances are predicted by the obtained function $g$.
\begin{algorithm}[tb!]  
	%\caption{HGKT: Construct Graph $G({\mathcal{V}}, {\mathcal{E}})$} 
    \caption{HGKT: Graph Construction $G({\mathcal{V}}, {\mathcal{E}})$}
	\label{alg:graph}
	\begin{algorithmic}[1]  
		\Require  
		The set of training data: $\left\{ D_{tr} \right\}$; the regularization penalty parameters $\lambda$; 
		%\Ensure
		%Graph $G({\mathcal{V}}, {\mathcal{E}})$;
		\State Classify the training data into $L$-parts, each part correspond to a class. For each class, connect the node with each other completely;
		\State Normalize these data into $n$-simplex space $\sum_n$ and compute the Wasserstein barycenter through (\ref{barycenter}) for each class to get the set of barycenters ${\mathcal{B}}$;
		\State For each class, take the node which is nearest to its barycenter as the representative node, get the set ${\mathcal{S}}$ ;
		\State For each representative node, select the $k$-nearest neighbors in ${\mathcal{S}}$ as its neighbors which are connected to the representative node; \\ \Return Adjacency list $G({\mathcal{V}}, {\mathcal{E}})$;
	\end{algorithmic} 
\end{algorithm}

\subsection{Heterogeneous Graph Embedding for ZSL}
%{\bfseries{Construct graph.}}
\subsubsection{Graph Construction.}
Inspired by the semi-supervised learning of GraphSAGE, we treat each instance as a node in the graph. For the instances that belong to the same class, they will be connected with each other. As a result, we can obtain $L$ complete subgraphs. Then the connections between these subgraphs are established. 
%In the training step of GNN, each single node will be affected by its neighboring nodes, for which we could draw on the idea of label propagation. %If two nodes in original space are similar, then they should be similar too in the embedding space, i.e., the relative distance between instances should be kept after embedding. Thus $k$-nearest neighbors technique is employed to connect these subgraphs.

Now we show our technique for how to select the representative nodes for each class. The node closest to the following obtained Wassersterin barycenter can be an effective means of representing the corresponding class.
% If the visual feature of each instance can be considered as a probability distribution in the feature space, the problem is transformed into the barycenter problem of probability distributions. Based on this observation, we introduce the Wasserstein distance as the difference measure between probabilities, and then search the Wasserstein barycenter of these probability distributions. 
Specifically for class $\ell$, given input visual feature probabilities $\{ x_k \}_{k=1}^{K_{\ell}}$, and penalty weights $\lambda^{\ell}\in \sum_{K_{\ell}}$, the Wasserstein barycenter $x_{\ell}$ is computed by $\mathcal{WB}(\lambda^{\ell}, \{ x_{k}\})$.

The representative node $s_{\ell}$ for each class can be selected the closest one to the Wasserstein barycenter for each class $\ell$:
\begin{equation}
    s_{\ell} = \arg\min_{x_i} \left\{ {\mathcal{L}}_C (x_{\ell}, x_i)\ \big|\ y_i = \ell \right\}.
\end{equation}The chosen representative nodes are connected by kNN, thus the whole neighbors of each representative node $s_{\ell}$ can be uniformly denoted as
\begin{equation}
    {{\cal{N}}(s_{\ell})} = {\cal{N}}_{\ell} (s_{\ell}) \cup {\cal{N}}_{r} (s_{\ell}),
\end{equation}
where ${\cal{N}}_{\ell}$ denotes the neighbors within each class $\ell$, and ${\cal{N}}_{r}$ denotes the neighbors chosen by using kNN in representative node set. Finally a heterogeneous graph is built, the detailed procedure of graph construction is described in Alg.~\ref{alg:graph}.
\begin{algorithm}[tb!]  
  \caption{HGKT: Training Procedure}  
  \label{alg:train}  
  \begin{algorithmic}[1]  
    \Require  
        Graph $G({\mathcal{V}}, {\mathcal{E}})$; Input seen class attributes $\left\{ p_v\right\}$, $\forall v \in {\mathcal{V}}$; Initialization $\{W_1, W_2\}$;%Aggregator functions $Mean$; Neighborhood function $N:v \rightarrow 2^{\mathcal{V}}$
    %\Ensure  
    %  Vector representations $\left\{ z_v \right\}$ for representative nodes;
    \State $h_v^0 \leftarrow p_v, \forall v \in {\mathcal{V}}$;
    \For{k = 1,2}
    \For{each $v \in {\mathcal{V}}$}
    \State $h_{{\cal{N}}(v)}^k \leftarrow {\rm Mean}(\{h_u^{k-1}, \forall u \in {\cal{N}}(v) \})$
    \State $h_v^k \leftarrow {\rm Relu}(W_k \cdot {\rm CONCAT}(h_v^{k-1}, \mu h_{{\cal{N}}(v)}^k)+ b)$
%    \State $h_{N(v)}^1 \leftarrow {\rm Mean}(\{h_u^1, \forall u \in N(v) \})$
%    \State $h_v^2 \leftarrow {\rm Relu}(W_2 \cdot {\rm CONCAT}(h_v^1, \mu h_{N(v)}^1)+ b)$
    \EndFor
    \EndFor
    \State $z_v \leftarrow z_v \cup h_v^2, \forall v \in {\rm representative \; nodes}$
    \\
    \Return Vector representations $\left\{ z_v\right\}$;  
  \end{algorithmic}
\end{algorithm}

\subsubsection{\textbf{Training Procedure.}}
After constructing the heterogeneous graph on training data, we further employ a two-layer GNN scheme for the training procedure. %The visual feature space is used for embedding. Specifically, given training data $D_{tr}$, in contrast to traditional approaches which usually map features to classes with function $f$, i.e., $y = f(x)$, we embed attributes to visual feature space with function $h$, i.e., $x = h(y)$. 
The $i$-th node has an attribute vector $p_i$ and visual feature $x_i$.
%The attributes vector was usually defined by human expert, while the visual feature was extracted by deep convolutional network, such as ResNet \cite{he2016deep}.

%To embed class protytope into features space, the attributes vector denotes the input is , and try to make output as close as possible to the visual feature of this node. 
The attribute vector is used as the initial embedding value for each node $v$. Then under the well-established structured graph, the unified neighbor embedding vector of each node $v$ can be obtained by averaging its neighbors' embedding vectors i.e.,
\begin{equation}
   h_{{\cal{N}}(v)} \leftarrow {\rm Mean}(\{h_u, \forall u \in {\cal{N}}(v) \}),
\end{equation}
where $h_{{\cal{N}}(v)}$ denotes the unified neighbor embedding vector of node $v$, $h_u$ denotes the embedding of node $u$.

After aggregation, the unified neighbor embedding vector is concatenated  with the original embedding vector of node $v$, followed by the embedding procedure:
%Compared with the embedding method in \cite{hamilton2017inductive}, a regularization parameter $\mu$ is introduced to balance the importance between unified neighbor embedding vector and original embedding vector, while some bias is added meanwhile. 
%The detailed iterative formulation is as follows:
\begin{equation}
    h_v \leftarrow \delta (W \cdot {\rm CONCAT}(h_v, \mu h_{{\cal{N}}(v)})+b),
\end{equation}where $h_v$ denotes the embedding vector of node $v$, $\delta$ denotes the activation function, $W$ denotes the neural network parameters and $b$ denotes the bias. Moreover, a regularization parameter $\mu$ is introduced to balance the importance between unified neighbor embedding vector and original embedding vector.
%Because the attributes are embedded into the visual feature space, it's natural to employ the least square loss other than traditional softmax or hinge loss.
We aim to minimize the distance between the embedding vector and visual feature vector as follows: %and learn these parameters by back propagation.
\begin{equation}
    {\rm loss} := \frac{1}{|D_{tr}|} \sum_{i=1}^{|D_{tr}|} \left\| h(y_i) - x_i \right\|^2 + \xi \| W \|^2,
\end{equation}where $\xi$ denotes the regularization parameter. During training, we can use seen class instances to estimate the parameters. The training procedure is depicted in Alg.~\ref{alg:train}.

 \begin{algorithm}[tb!]  
  \caption{HGKT: Testing Procedure}  
  \label{alg:test}  
  \begin{algorithmic}[1]  
    \Require  
        Graph $G({\mathcal{V}}, {\mathcal{E}})$; All seen class embedding $\left\{z_{v}\right\}$; Input unseen class attributes {$\left\{\tilde{p}_{u}\right\}$}; test visual feature $x_t$;
    %\Ensure  
    %  The predict class $y_t$ for test visual feature $x_t$;
    \State Compute the $k$-nearest seen classes for each unseen class and converge them into ${\cal{N}}(\tilde{p}_{u})$;
    \State Get the embedding result $\left\{ \tilde{z}_{u}\right\}$ by using GNN;
    %\State Get the embedding for all classes $z_v = z_v \cup z_{uv}$;
    \State For the test visual feature $x_t$, choose the nearest embedding's label within $\left\{ z_{v}\right\} \cup \left\{\tilde{z}_{u}\right\}$ as prediction $y_t$;  \\
    \Return predict result $y_t$
  \end{algorithmic}
\end{algorithm} 

%\smallskip
\subsubsection{Testing Procedure.}
%In testing, we need to predict the embedding vector for all classes.
For seen classes, the embedding vectors have been computed in the train phase.
As for new unseen classes, we need to calculate and select $k$ nearest seen classes as its neighbors, %Further we create a new node in the heterogeneous graph for this unseen class and connect it with the representative nodes of the above chosen $k$-nearest seen classes, 
while their embedding vectors can be aggregated directly through the graph and the constructed GNN model. GNN is employed to maintain the topology between classes, as result the relative location of the unseen class can be determined by its neighbor classes. Finally, the visual feature of test samples (could from both seen and unseen classes) are compared with all seen and unseen class embedding vectors, and select the nearest one as its prediction. The detailed procedure of testing is shown in Alg.~\ref{alg:test}.

\section{Experiments}
%{\bfseries{Datasets.}}
\subsection{Protocols}
\subsubsection{Datasets.}
We select five public benchmark datasets for evaluation, i.e., SUN \cite{patterson2012sun}, CUB \cite{wah2011caltech}, AwA1 \cite{lampert2014attribute}, AwA2 \cite{xian2018zero} and aPY \cite{farhadi2009describing}. The detailed setting of the datasets are summarized in Table \ref{tab:statistics}.
%\vspace{-0.4cm}

%apy ok

%\smallskip
%\noindent {\bfseries{Setting.}} 
\subsubsection{Setting.}
The same deep features extracted from ResNet-101~\cite{he2016deep} and the same attributes presented in \cite{xian2018zero} are employed in this experiment. The code is implemented in Python and the network is built by Pytorch. The dimension of intermediate layers of our two-layer GNN for five benchmarks are all set to be $1600$. The activation layer uses the $leakyRelu$ function with slope of $0.2$, and {\sc{Adam}} optimizer \cite{kingma2014adam} with a learning rate $1e^{-4}$ and a weight decay of $0.001$ is adopted for training. The regularization parameter $\mu$ are selected in $\{0.01, 0.1, 0.5, 1\}$ by validation set. For simplicity, we set the number of neighbors for constructing graph equals to the number of neighbors for unseen classes, and then select $k$ in $\{1,2,5,10,50\}$. In experiments, the value $k=2$ comes up as the best setting. 
%It is noting that for the $SUN$ dataset, normalize the features before training can get higher accuracy.

Moreover, for fairness, we employ the proposed splits (PS) \cite{xian2018zero} to evaluate the average-class top-1 accuracy on unseen classes $\mathcal{Y}^{ts}$ and seen classes $\mathcal{Y}^{tr}$. In addition the harmonic mean $H$ of training and test accuracies is used to measure the comprehensive performance of different methods, for both seen and unseen data. 

%Because some algorithms can get high performance on seen classes, while very low performance on unseen classes, however good algorithm should achieve good performance on both seen and unseen classes.
\begin{equation}
    H = \frac{2*acc_{\mathcal{Y}^{tr}}*acc_{\mathcal{Y}^{ts}}}{acc_{\mathcal{Y}^{tr}} + acc_{\mathcal{Y}^{ts}}}.
\end{equation}

%\smallskip
%\noindent {\bfseries{Competing Methods}}
\subsubsection{Competing Methods.}
State-of-the-art methods are selected to compare with our method.
Note that \cite{Long2017Zero} and \cite{zhu2019generalized} are both agnostic to unseen data, being similar to our approach. However, they use original images instead of extracted features proposed in \cite{xian2018zero} for model training, the experimental setup of these two algorithms is different from ours. Moreover, the method in \cite{zhu2019generalized} need to decompose an image into visual parts, but images in some datasets like SUN cannot be easily decomposed as they are often compact and consistent across different scenes. Thus in the corresponding experiment, we do not take these two methods into account.
%\smallskip

\begin{table*}[tb!]
	\caption{Comparison of GZSL methods on public datasets. Measuring $ts$ = Top-1 Accuracy on $\mathcal{Y}^{ts}$, $tr$ = Top-1 Accuracy on $\mathcal{Y}^{tr}$, $H$ = Harmonic Mean. The result of the compared methods are taken from \cite{xian2018zero,zhang2017learning,annadani2018preserving}. Best results are marked in bold.}
%b	\vspace{-10pt}
	\resizebox{\textwidth}{!}{
		\begin{tabular}{rcccccccccccccccc}
		\toprule
			Dataset& \multicolumn{3}{c}{SUN} & \multicolumn{3}{c}{CUB} & \multicolumn{3}{c}{AwA1} & \multicolumn{3}{c}{AwA2} & \multicolumn{3}{c}{aPY} \\
			\midrule
			\multicolumn{1}{r|}{Method} & $ts$ & $tr$ & \multicolumn{1}{c|}{$H$} & $ts$ & $tr$ & \multicolumn{1}{c|}{$H$} & $ts$ & $tr$ & \multicolumn{1}{c|}{$H$} & $ts$ & $tr$ & \multicolumn{1}{c|}{$H$} & $ts$ & $tr$ & $H$ \\ 
				\midrule
			\multicolumn{1}{r|}{DAP~\cite{lampert2014attribute}} & 4.2 & 25.1 & \multicolumn{1}{c|}{7.2} & 1.7 & 67.9 & \multicolumn{1}{c|}{3.3} & 0.0 & \textbf{88.7} & \multicolumn{1}{c|}{0.0} & 0.0 & 84.7 & \multicolumn{1}{c|}{0.0} & 4.8 & 78.3 & 9.0 \\
			\multicolumn{1}{r|}{IAP~\cite{lampert2014attribute}} & 1.0 & 37.8 & \multicolumn{1}{c|}{1.8} & 0.2 & 72.8 & \multicolumn{1}{c|}{0.4} & 2.1 & 78.2 & \multicolumn{1}{c|}{4.1} & 0.9 & 87.6 & \multicolumn{1}{c|}{1.8} & 5.7 & 65.6 & 10.4 \\
			\multicolumn{1}{r|}{CONSE~\cite{norouzi2013zero}} & 6.8 & 39.9 & \multicolumn{1}{c|}{11.6} & 1.6 & 72.2 & \multicolumn{1}{c|}{3.1} & 0.4 & 88.6 & \multicolumn{1}{c|}{0.8} & 0.5 & \textbf{90.6} & \multicolumn{1}{c|}{1.0} & 0.0 & \textbf{91.2} & 0.0 \\
			\multicolumn{1}{r|}{CMT~\cite{socher2013zero}} & 8.1 & 21.8 & \multicolumn{1}{c|}{11.8} & 7.2 & 49.8 & \multicolumn{1}{c|}{12.6} & 0.9 & 87.6 & \multicolumn{1}{c|}{1.8} & 0.5 & 90.0 & \multicolumn{1}{c|}{1.0} & 1.4 & 85.2 & 2.8 \\
			\multicolumn{1}{r|}{CMT*~\cite{socher2013zero}} & 8.7 & 28.0 & \multicolumn{1}{c|}{13.3} & 4.7 & 60.1 & \multicolumn{1}{c|}{8.7} & 8.4 & 86.9 & \multicolumn{1}{c|}{15.3} & 8.7 & 89.0 & \multicolumn{1}{c|}{15.9} & 10.9 & 74.2 & 19.0 \\
			\multicolumn{1}{r|}{SSE~\cite{zhang2015zero}} & 2.1 & 36.4 & \multicolumn{1}{c|}{4.0} & 8.5 & 46.9 & \multicolumn{1}{c|}{14.4} & 7.0 & 80.5 & \multicolumn{1}{c|}{12.9} & 8.1 & 82.5 & \multicolumn{1}{c|}{14.8} & 0.2 & 78.9 & 0.4 \\
			\multicolumn{1}{r|}{LATEM\cite{xian2016latent}} & 14.7 & 28.8 & \multicolumn{1}{c|}{19.5} & 15.2 & 57.3 & \multicolumn{1}{c|}{24.0} & 7.3 & 71.7 & \multicolumn{1}{c|}{13.3} & 11.5 & 77.3 & \multicolumn{1}{c|}{20.0} & 0.1 & 73.0 & 0.2 \\
			\multicolumn{1}{r|}{ALE~\cite{akata2016label}} & 21.8 & 33.1 & \multicolumn{1}{c|}{26.3} & 23.7 & 62.8 & \multicolumn{1}{c|}{34.4} & 16.8 & 76.1 & \multicolumn{1}{c|}{27.5} & 14.0 & 81.8 & \multicolumn{1}{c|}{23.9} & 4.6 & 73.7 & 8.7 \\
			\multicolumn{1}{r|}{DEVISE~\cite{frome2013devise}} & 16.9 & 27.4 & \multicolumn{1}{c|}{20.9} & 23.8 & 53.0 & \multicolumn{1}{c|}{32.8} & 13.4 & 68.7 & \multicolumn{1}{c|}{22.4} & 17.1 & 74.7 & \multicolumn{1}{c|}{27.8} & 4.9 & 76.9 & 9.2 \\
			\multicolumn{1}{r|}{SJE~\cite{akata2015evaluation}} & 14.7 & 30.5 & \multicolumn{1}{c|}{19.8} & 23.5 & 59.2 & \multicolumn{1}{c|}{33.6} & 11.3 & 74.6 & \multicolumn{1}{c|}{19.6} & 8.0 & 73.9 & \multicolumn{1}{c|}{14.4} & 3.7 & 55.7 & 6.9 \\
			\multicolumn{1}{r|}{ESZSL~\cite{romera2015embarrassingly}} & 11.0 & 27.9 & \multicolumn{1}{c|}{15.8} & 12.6 & 63.8 & \multicolumn{1}{c|}{21.0} & 6.6 & 75.6 & \multicolumn{1}{c|}{12.1} & 5.9 & 77.8 & \multicolumn{1}{c|}{11.0} & 2.4 & 70.1 & 4.6 \\
			\multicolumn{1}{r|}{SYNC~\cite{changpinyo2016synthesized}} & 7.9 & \textbf{43.4} & \multicolumn{1}{c|}{13.4} & 11.5 & \textbf{70.9} & \multicolumn{1}{c|}{19.8} & 8.9 & 87.3 & \multicolumn{1}{c|}{16.2} & 10.0 & 90.5 & \multicolumn{1}{c|}{18.0} & 7.4 & 66.3 & 13.3 \\
			\multicolumn{1}{r|}{SAE~\cite{kodirov2017semantic}} & 8.8 & 18.0 & \multicolumn{1}{c|}{11.8} & 7.8 & 54.0 & \multicolumn{1}{c|}{13.6} & 1.8 & 77.1 & \multicolumn{1}{c|}{3.5} & 1.1 & 82.2 & \multicolumn{1}{c|}{2.2} & 0.4 & 80.9 & 0.9 \\
			\multicolumn{1}{r|}{GFZSL~\cite{verma2017simple}} & 0.0 & 39.6 & \multicolumn{1}{c|}{0.0} & 0.0 & 45.7 & \multicolumn{1}{c|}{0.0} & 1.8 & 80.3 & \multicolumn{1}{c|}{3.5} & 2.5 & 80.1 & \multicolumn{1}{c|}{4.8} & 0.0 & 83.3 & 0.0 \\
			\multicolumn{1}{r|}{DEM~\cite{zhang2017learning}} & 20.5 & 34.3 & \multicolumn{1}{c|}{25.6} & 19.6 & 57.9 & \multicolumn{1}{c|}{29.2} & 32.8 & {84.7} & \multicolumn{1}{c|}{47.3} & 30.5 & {86.4} & \multicolumn{1}{c|}{45.1} & 11.1 & 75.1 & 19.4 \\
			\multicolumn{1}{r|}{PSRZSL~\cite{annadani2018preserving}} & 20.8 & 37.2 & \multicolumn{1}{c|}{26.7} & 24.6 & 54.3 & \multicolumn{1}{c|}{33.9} & - & - & \multicolumn{1}{c|}{-} & 20.7 & 73.8 & \multicolumn{1}{c|}{32.3} & 13.5 & 51.4 & 21.4 \\ \hline
		%	\multicolumn{1}{c}{PSRZSL\cite{annadani2018preserving}} & \multicolumn{1}{c}{20.8} & \multicolumn{1}{c}{37.2} & \multicolumn{1}{c|}{26.7} & \multicolumn{1}{c}{\textbf{24.6}} & \multicolumn{1}{c}{54.3} & \multicolumn{1}{c|}{33.9} & - & - & \multicolumn{1}{c|}{-} & \multicolumn{1}{c}{20.7} & \multicolumn{1}{c}{73.8} & \multicolumn{1}{c|}{32.3} & \multicolumn{1}{c}{13.5} & \multicolumn{1}{c}{51.4} & \multicolumn{1}{c}{21.4} \\ \hline
			\multicolumn{1}{r|}{HGKT (Ours)} & {\textbf{22.3}} & {36.5} & \multicolumn{1}{c|}{\textbf{27.7}} & {\textbf{25.2}} & {56.9} & \multicolumn{1}{c|}{{\textbf{34.9}}} & \textbf{39.4} & 83.5 & \multicolumn{1}{c|}{\textbf{53.6}} & \textbf{37.9} & 86.5 & \multicolumn{1}{c|}{\textbf{52.7}} & \textbf{18.3} & {79.0} & \textbf{29.7} \\ 
			\bottomrule
	\end{tabular}}
	\label{Main-results}
\end{table*}

\begin{table}[tb!]
    \centering
	\caption{Benchmark datasets. $\mathcal{Y}^{tr}$ and $\mathcal{Y}^{ts}$ represent the seen class and the unseen class respectively.}
	%\vspace{-10pt}
	\resizebox{0.6\textwidth}{!}{
		\begin{tabular}{c|ccc|cc|cc}
		\hline
			\multicolumn{4}{c|}{Class Number}                        & \multicolumn{4}{c}{Number of Images}                                  \\ \hline
			&           &                &                  & \multicolumn{2}{c|}{Training Set}  & \multicolumn{2}{c}{Test Set}      \\ \hline
			Dataset & Attribute & $\mathcal{Y}^{tr}$ & $\mathcal{Y}^{ts}$ & $\mathcal{Y}^{tr}$ & $\mathcal{Y}^{ts}$ & $\mathcal{Y}^{tr}$ & $\mathcal{Y}^{ts}$ \\ \hline
			SUN     & 102       & 645            & 72               & 10320          & 0                & 2580           & 1440             \\
			CUB     & 312       & 150            & 50               & 7057           & 0                & 1764           & 2967             \\
			AwA1 & 85        & 40             & 10               & 19832         & 0                & 4958           & 5685             \\
			AwA2 & 85        & 40             & 10               & 23527           & 0                & 5882           & 7913             \\
			aPY     & 64        & 20             & 12               & 5932           & 0                & 1483           & 7924             \\ \hline
	\end{tabular}}
	\label{tab:statistics}
\end{table}

\begin{table}[tb!]
\centering
\caption{Average per-class accuracy for conventional ZSL.}
%\resizebox{0.8\textwidth}{!}{
    \begin{tabular}{|c|ccccc|}
    \toprule
    Method & SUN & CUB & AwA1 & AwA2 & aPY \\ \midrule
    DAP~\cite{lampert2014attribute} & 39.9 & 40.0 & 44.1 & 46.1 & 33.8 \\
    IAP~\cite{lampert2014attribute} & 19.4 & 24.0 & 35.9 & 35.9 & 36.6 \\
    CONSE~\cite{norouzi2013zero} & 38.8 & 34.3 & 45.6 & 44.5 & 26.9 \\
    CMT~\cite{socher2013zero} & 39.9 & 34.6 & 39.5 & 37.9 & 28.0 \\
    SSE~\cite{zhang2015zero} & 51.5 & 43.9 & 60.1 & 61.0 & 34.0 \\
    LATEM~\cite{xian2016latent} & 55.3 & 49.3 & 55.1 & 55.8 & 35.2 \\
    ALE~\cite{akata2016label} & 58.1 & 54.9 & 59.9 & 62.5 & 39.7 \\
    DEVISE~\cite{frome2013devise} & 56.5 & 52.0 & 54.2 & 59.7 & 39.8 \\
    SJE~\cite{akata2015evaluation} & 53.7 & 53.9 & 65.6 & 61.9 & 32.9 \\
    ESZSL~\cite{romera2015embarrassingly} & 54.5 & 53.9 & 58.2 & 58.6 & 38.3 \\
    SYNC~\cite{changpinyo2016synthesized} & 56.3 & 55.6 & 54.0 & 46.6 & 23.9 \\
    SAE~\cite{kodirov2017semantic} & 40.3 & 33.3 & 53.0 & 54.1 & 8.3 \\
    GFZSL~\cite{verma2017simple} & 60.6 & 49.3 & 68.3 & 63.8 & 38.4 \\
    DEM~\cite{zhang2017learning} & 61.9 & 50.1 & 64.2 & 61.9 & 34.5 \\
    PSRZSL~\cite{annadani2018preserving} & 61.4 & \textbf{56.0} & - & 63.8 & 38.4 \\ \hline
    HGKT (ours) & \textbf{62.1} & 54.2 & \textbf{69.1} & \textbf{68.9} & \textbf{41.5} \\ \bottomrule
\end{tabular}
%}
\label{Con-results}
\end{table}

%\noindent {\bfseries{Results and Analysis.}}
\subsection{Results and Discussion}
\subsubsection{Overall Comparison with Peer Methods}
We first make a comparison with state-of-the-arts for conventional ZSL which only consider unseen classes at test time. The results are listed in Table \ref{Con-results}. Compared to the baseline DEM, our proposed method outperforms in all benchmark datasets. Moreover, we achieve the state-of-the-art in SUN, AwA1, AwA2, aPY datasets. In particular, our result is further improved to 68.9\% in AwA2 dataset, which is 5.1\% higher that the best result reported so far. Despite we obtain 54.2\% on CUB dataset which is sightly less than PSRZSL, our approach HGKT performs much better on the generalized zero-shot setting, as illustrated next.

\begin{figure}[tb!]
	\centering
	\includegraphics[width=0.45\textwidth]{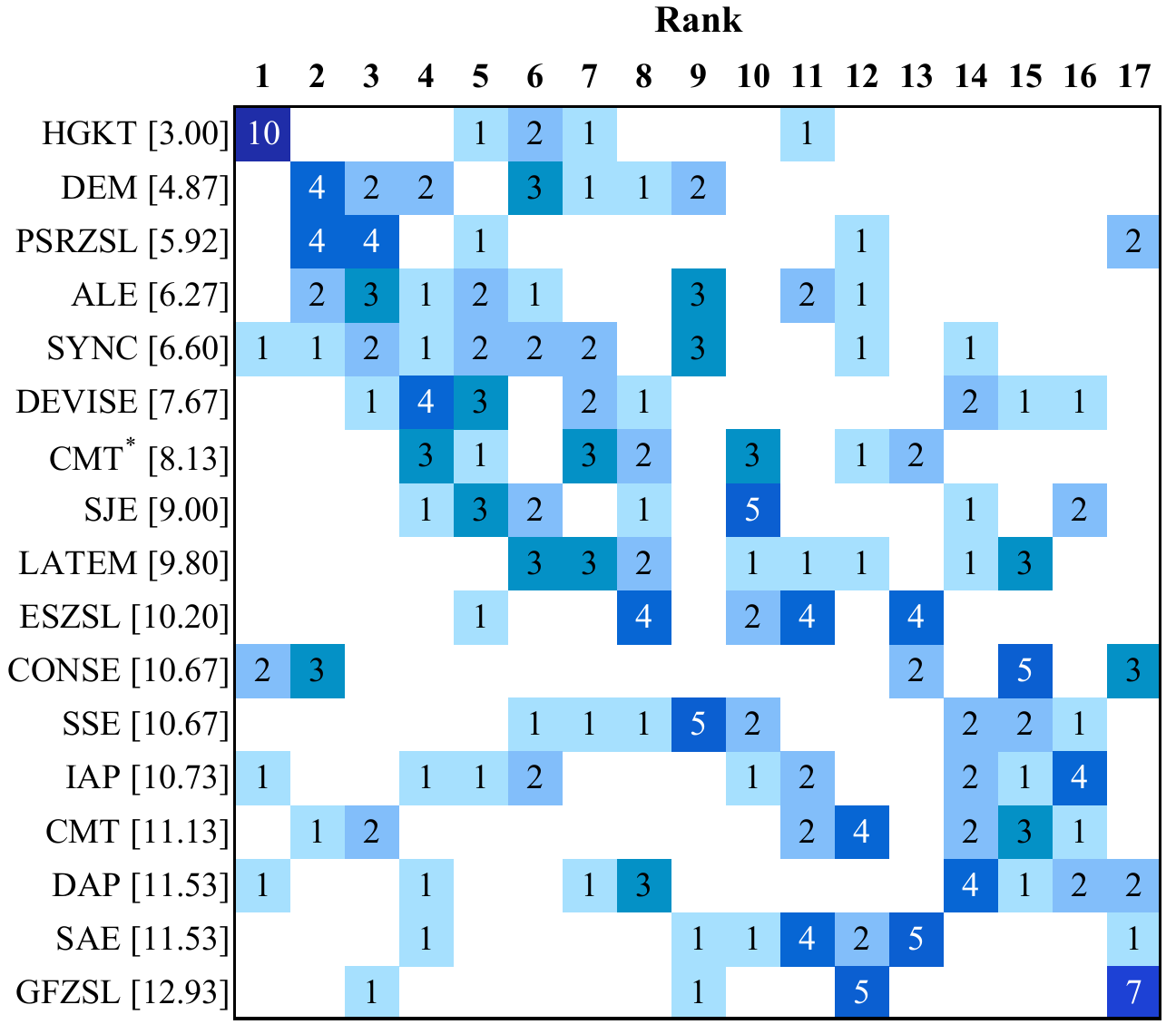}
	\vspace{-10pt}
	\caption{Statistic about the value of $ts$, $tr$ and $H$ in Table \ref{Main-results}. Element $(i,j)$ indicates number of times model $i$ ranks at $j$th over all $5 \times 3$ columns. Models are ordered by their mean rank (displayed in brackets).}
	\label{fig:rank}
\end{figure}

For GZSL, the evaluation results are listed in Table \ref{Main-results}.
Note that our approach obtains the highest top-$1$ accuracy of $H$ and unseen classes on all datasets. For AwA1, AwA2 and aPY datasets, our method improves the top-$1$ accuracy by a high margin. Especially, compared to the key baseline DEM method, our method gains $10.3\%$, $6.3\%$ and $7.6\%$ improvement on aPY, AwA1 and AwA2 datasets respectively.
Compared to the state-of-art approach PSRZSL~\cite{annadani2018preserving}, which relies on using semantic relations to learn the embedding, our algorithm can improve prominently $20.4\%$ on AwA2 datasets. It can be seen that our method improves the accuracy of unseen classes notably, which usually dominates the harmonic mean $H$. Moreover, we also count Table \ref{Main-results} about seen classes, unseen classes and $H$ in Fig. \ref{fig:rank}, which further illustrates the effectiveness of our approach.

\begin{figure}[tb!]
	\centering
	\includegraphics[width=0.45\textwidth]{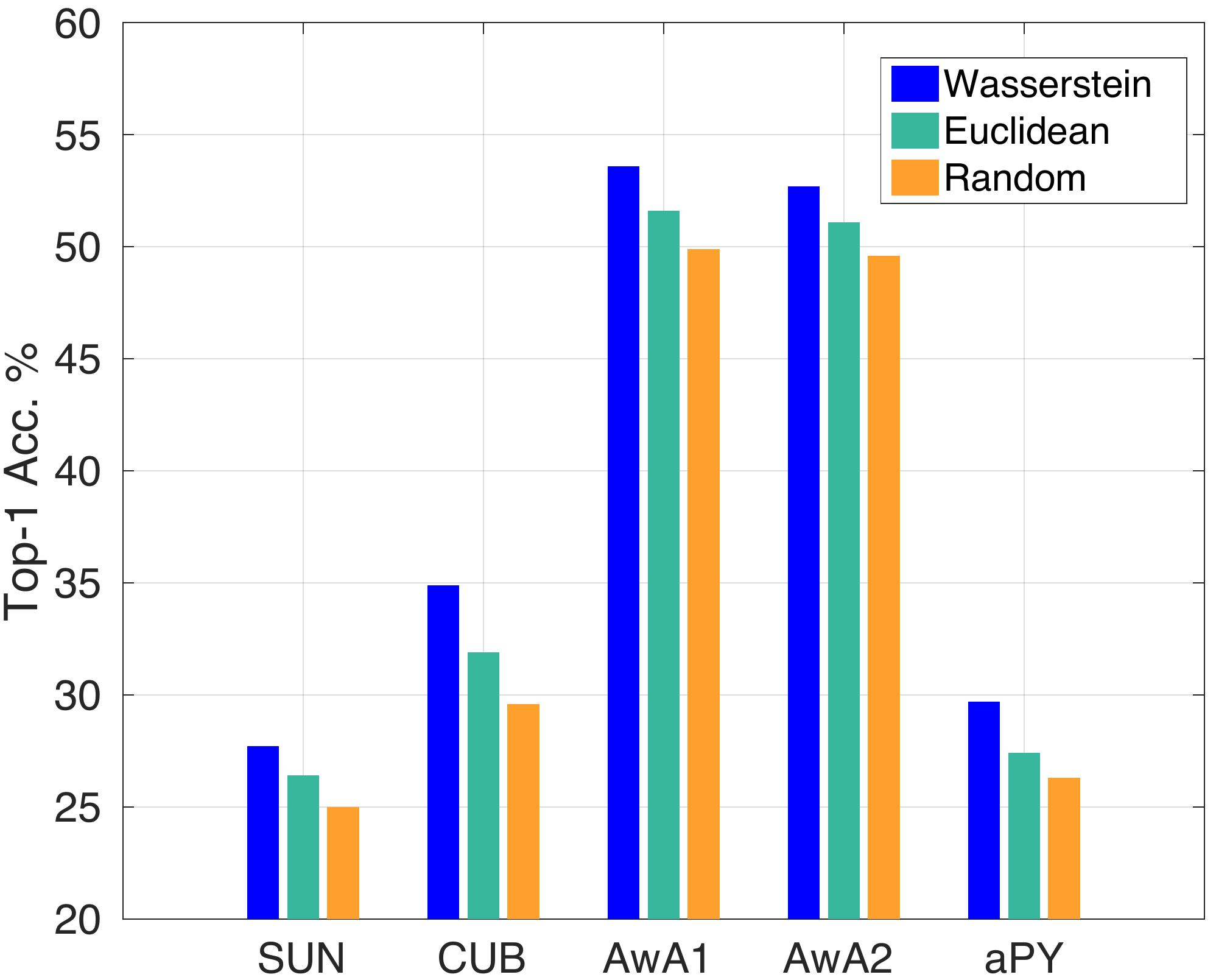}
	%\vspace{-10pt}
	\caption{Harmonic mean of three node selecting methods.}
	\label{fig:acc}
	\vspace{-10pt}
\end{figure}

\begin{figure*}[tb!]
	\centering
	\subfigure{\includegraphics[width=0.44\textwidth]{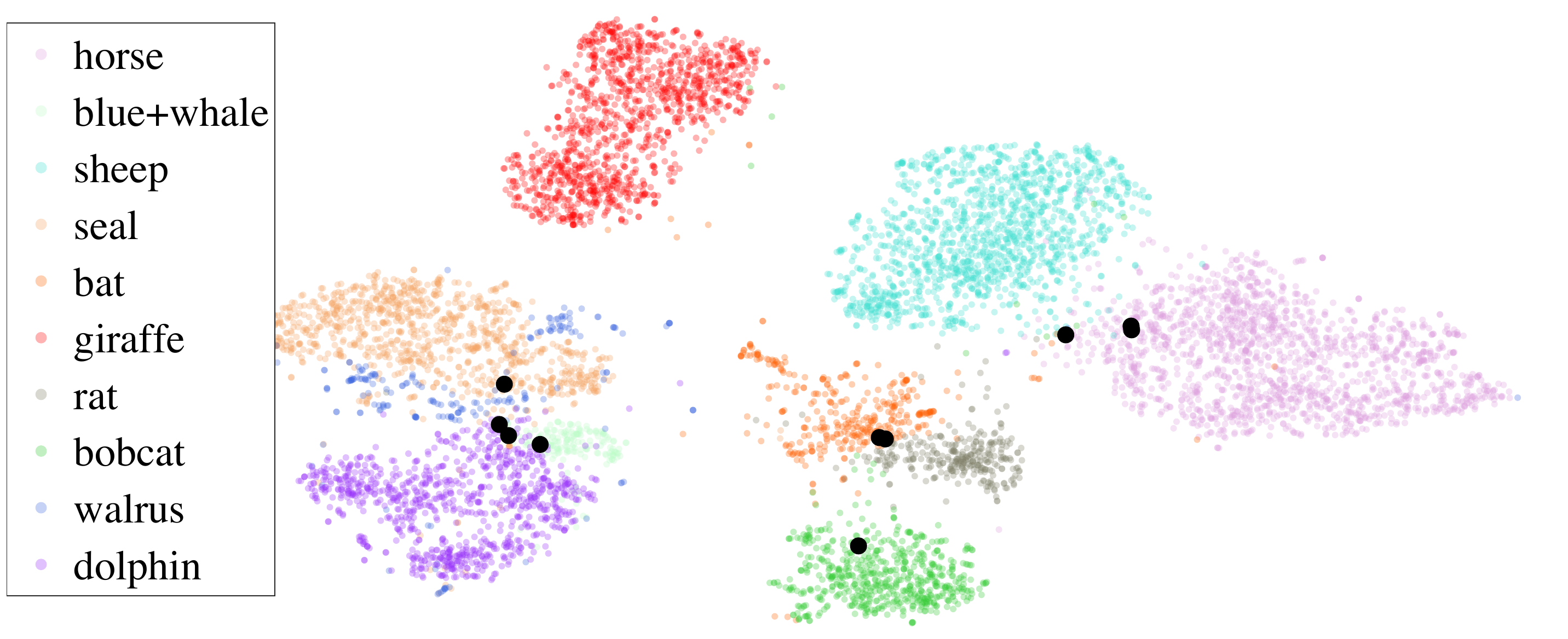}} %\vspace{-10pt}
	\subfigure{\includegraphics[height=0.2\textwidth]{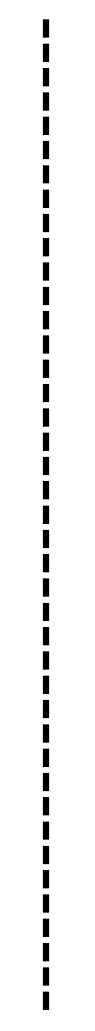}} %\vspace{-10pt}
	\subfigure{\includegraphics[width=0.36\textwidth]{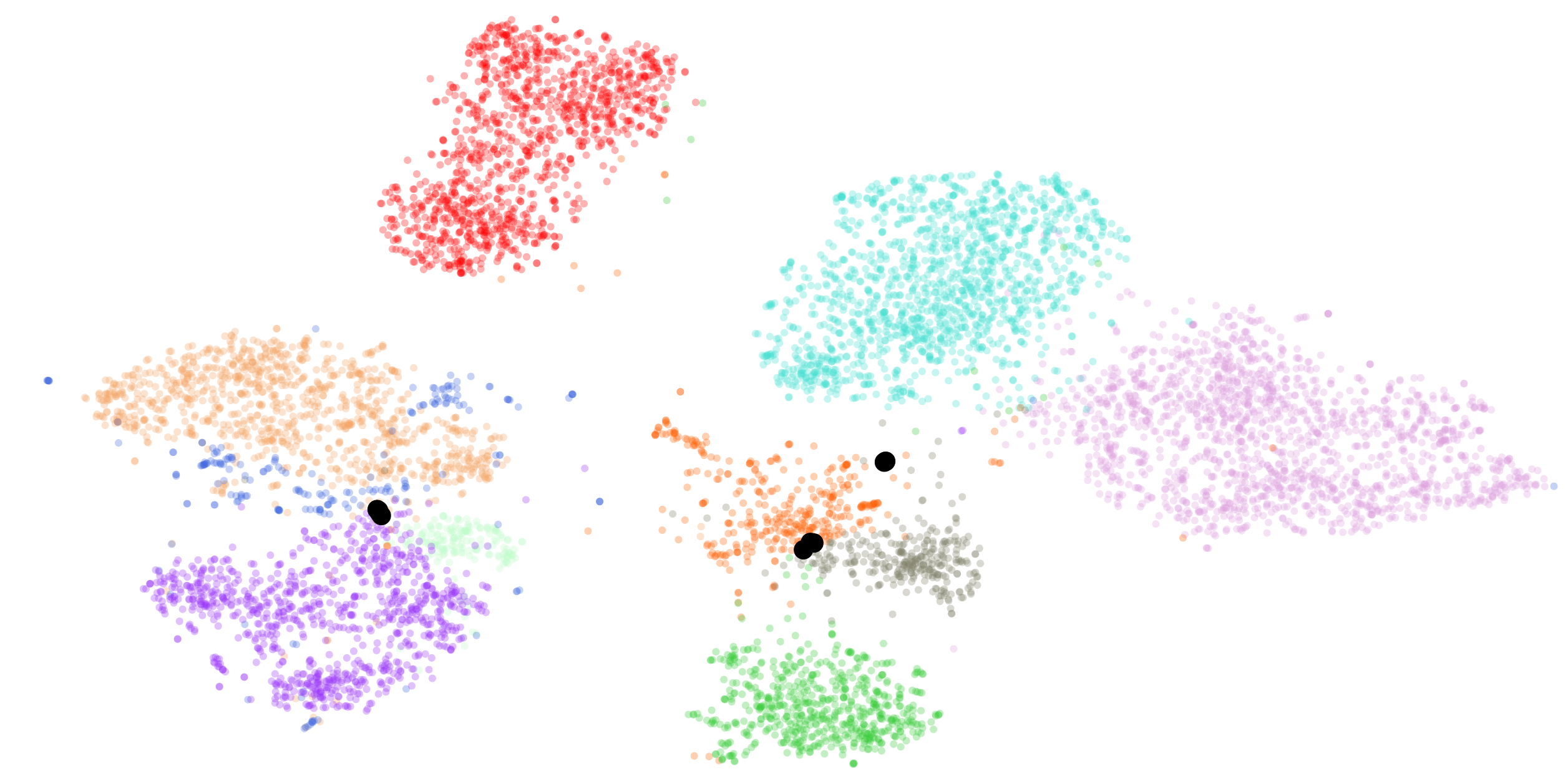}} %\vspace{-10pt}
	\caption{T-SNE visualization of the distribution of the ten unseen classes on AwA2. Black dots denote the embedding vectors for unseen classes obtained by different methods. Left: embedding vectors by our method; right: by the baseline DEM.}
	\label{fig:emb}
\end{figure*}
\begin{figure*}[tb!]
	\centering
	\includegraphics[width=0.8\textwidth]{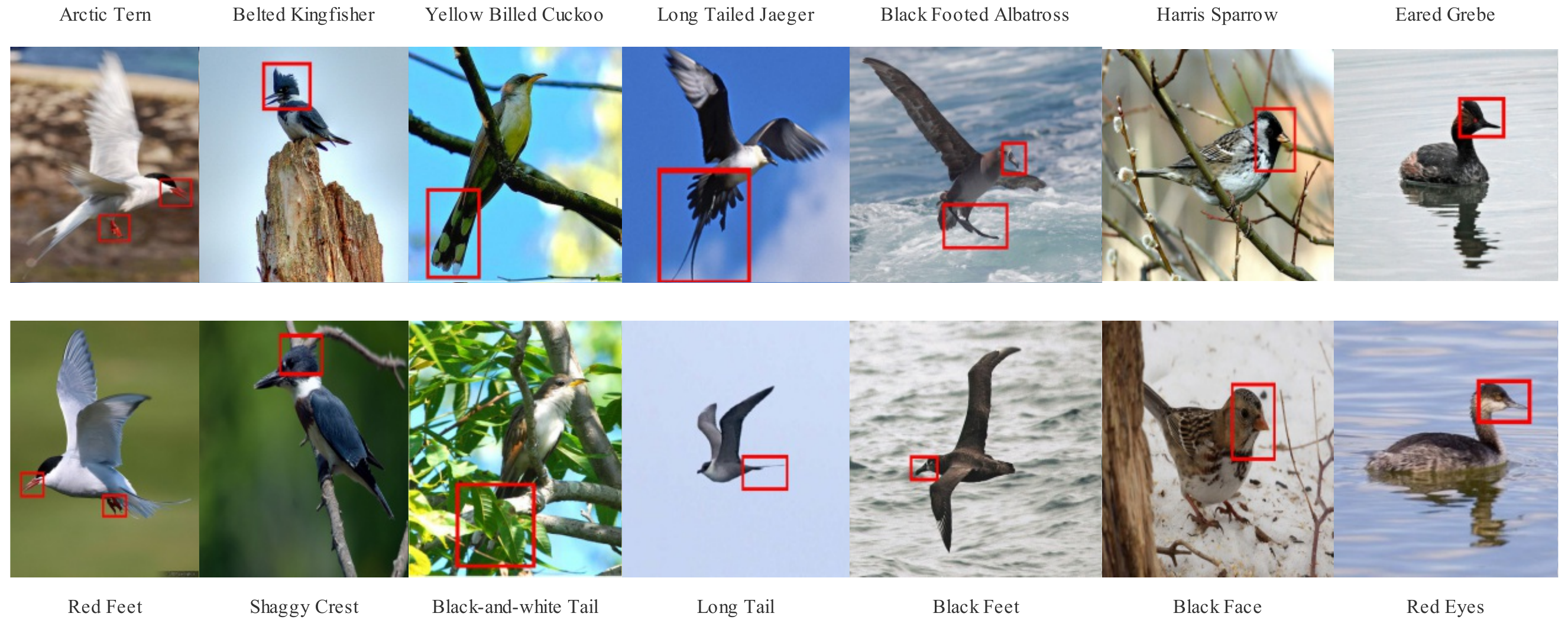}
	\vspace{-10pt}
	\caption{Representative images selected by Wasserstein barycenter (top) and Euclidean barycenter (bottom) on CUB dataset.}
	\label{fig:animals}
\end{figure*}
\begin{figure*}[tb!]
	\centering
	\subfigure[SUN]{\includegraphics[width=0.175\textwidth]{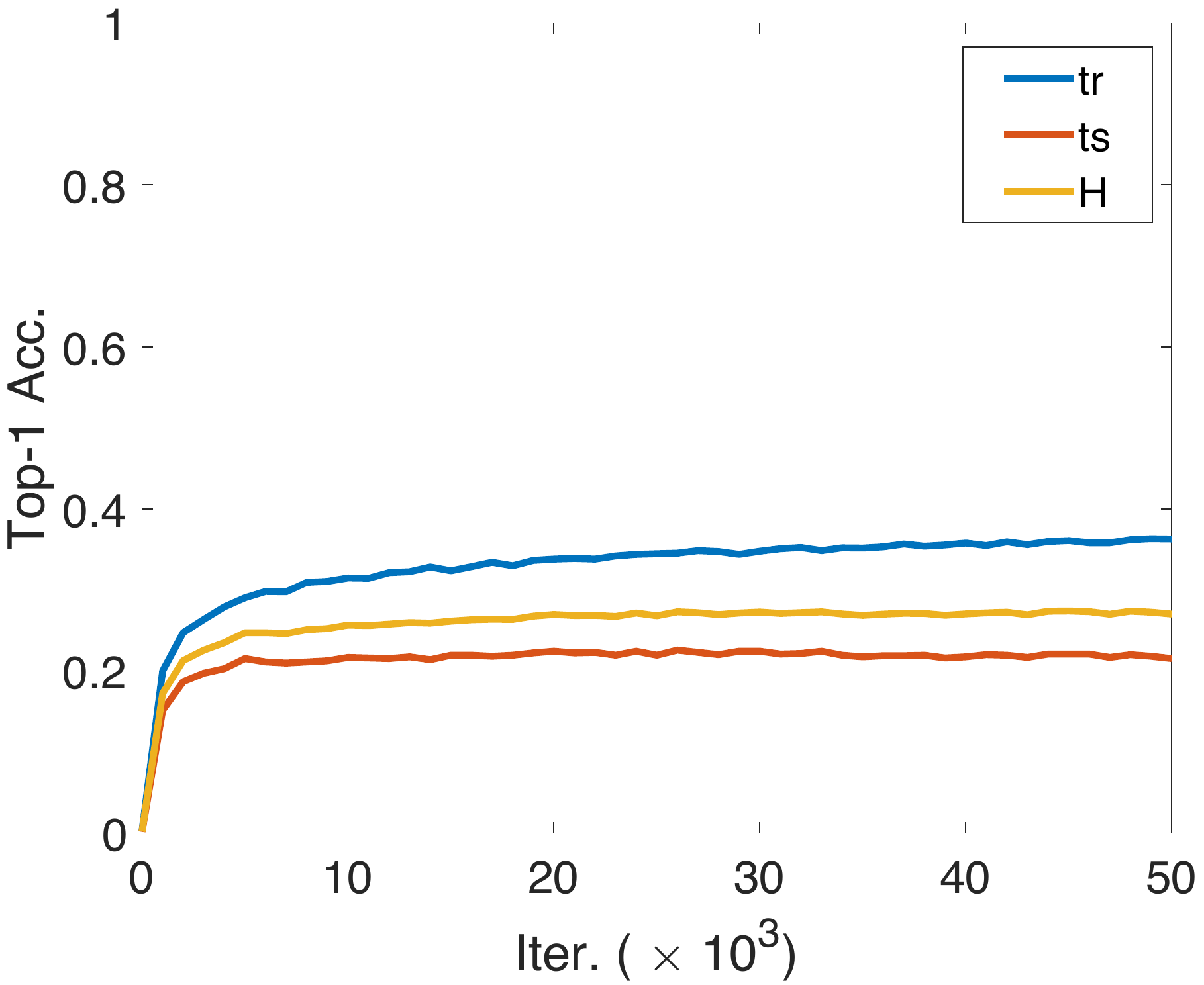}} %\vspace{-10pt}
	\subfigure[CUB]{\includegraphics[width=0.175\textwidth]{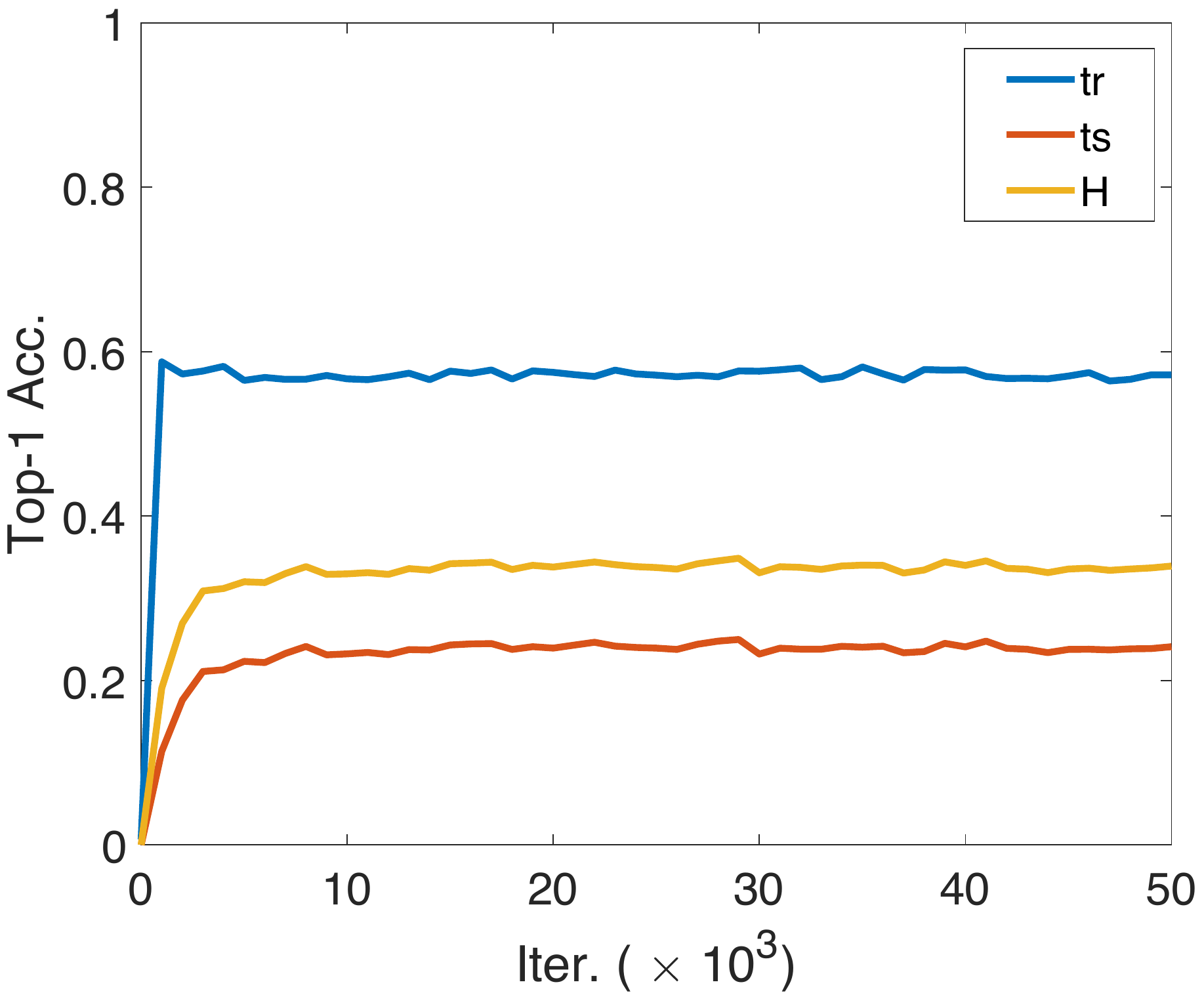}}%\\ \vspace{-10pt}
	\subfigure[AwA1]{\includegraphics[width=0.175\textwidth]{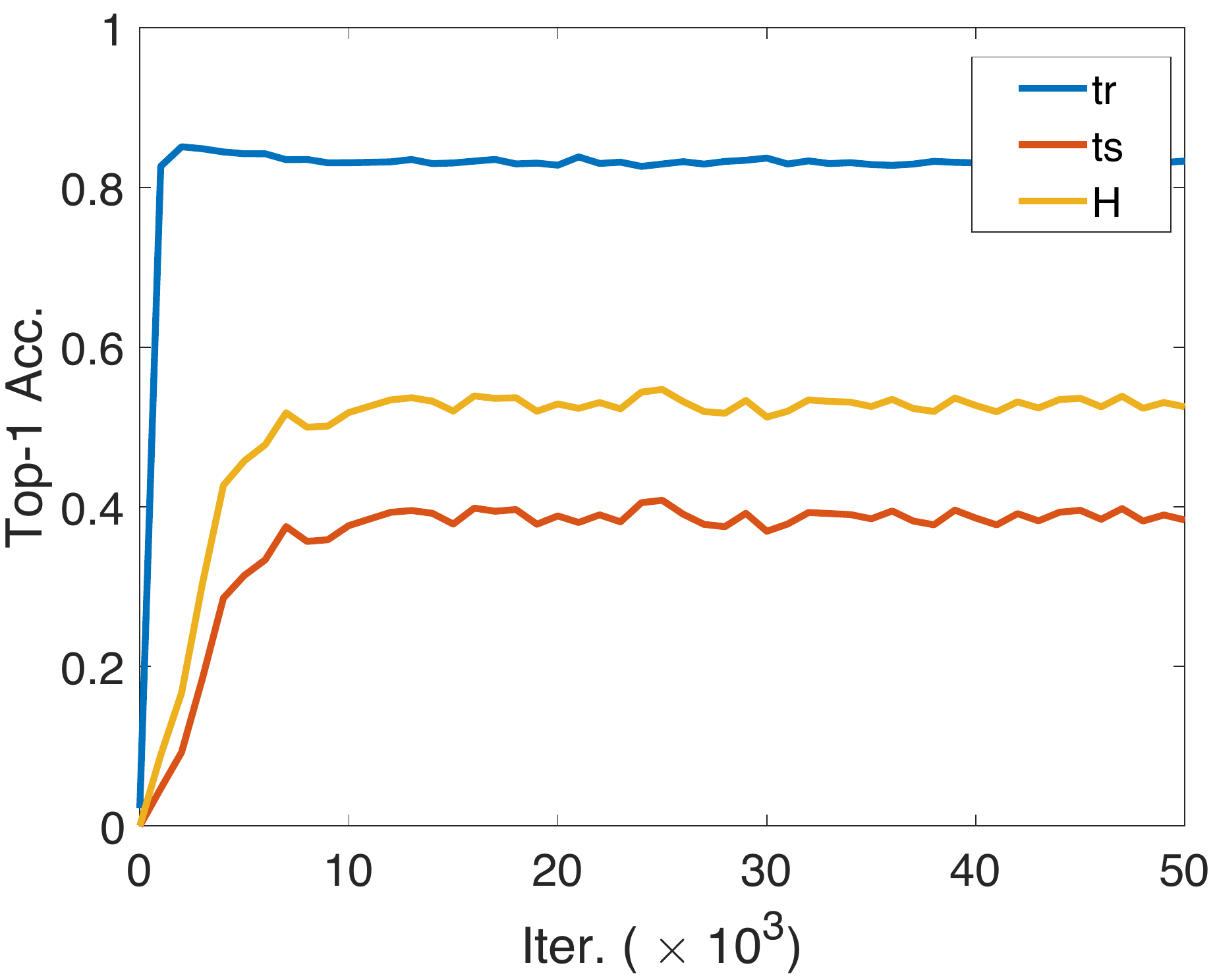}} %\vspace{-10pt}
	\subfigure[AwA2]{\includegraphics[width=0.175\textwidth]{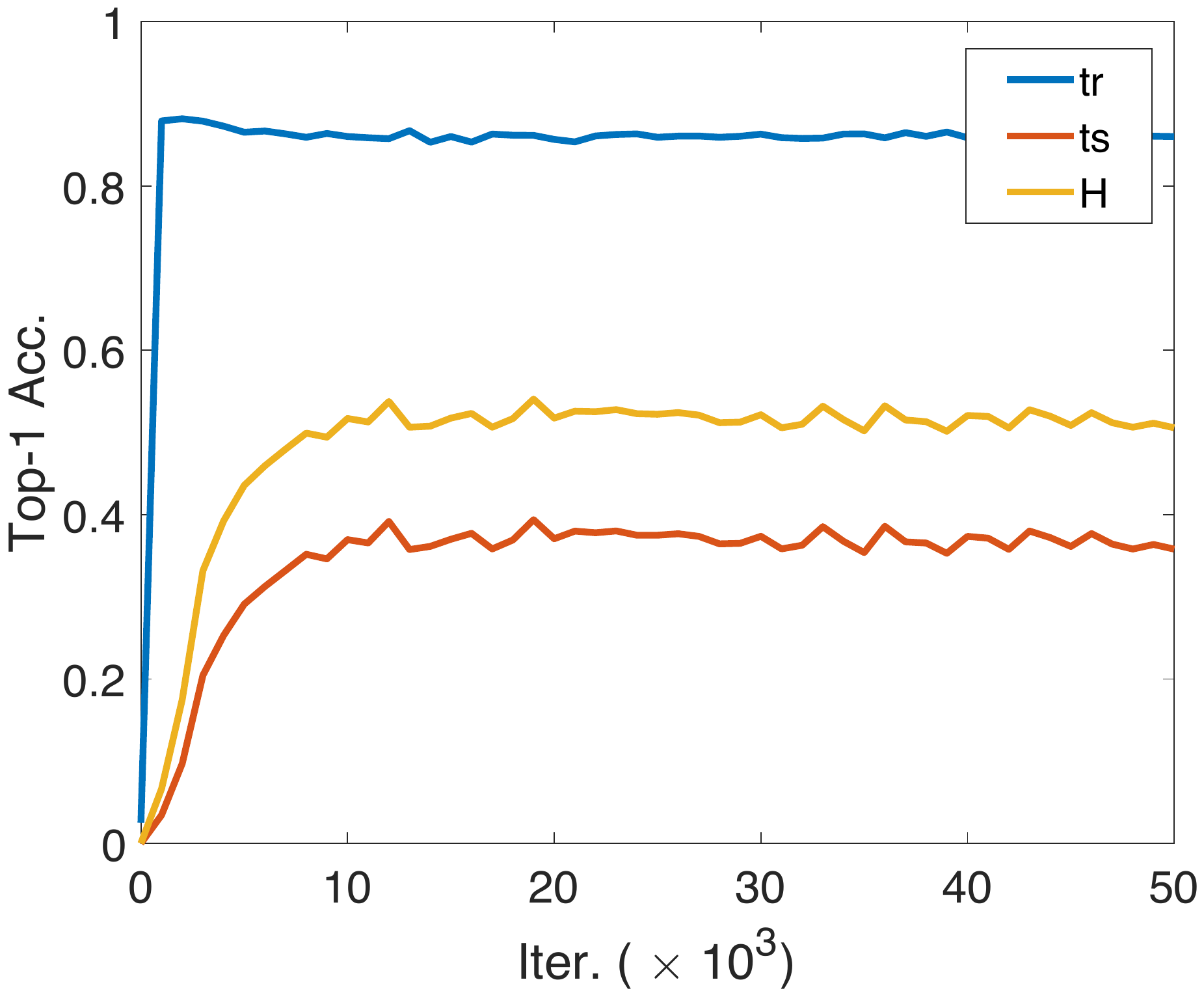}}%\\ \vspace{-10pt}
	\subfigure[aPY]{\includegraphics[width=0.175\textwidth]{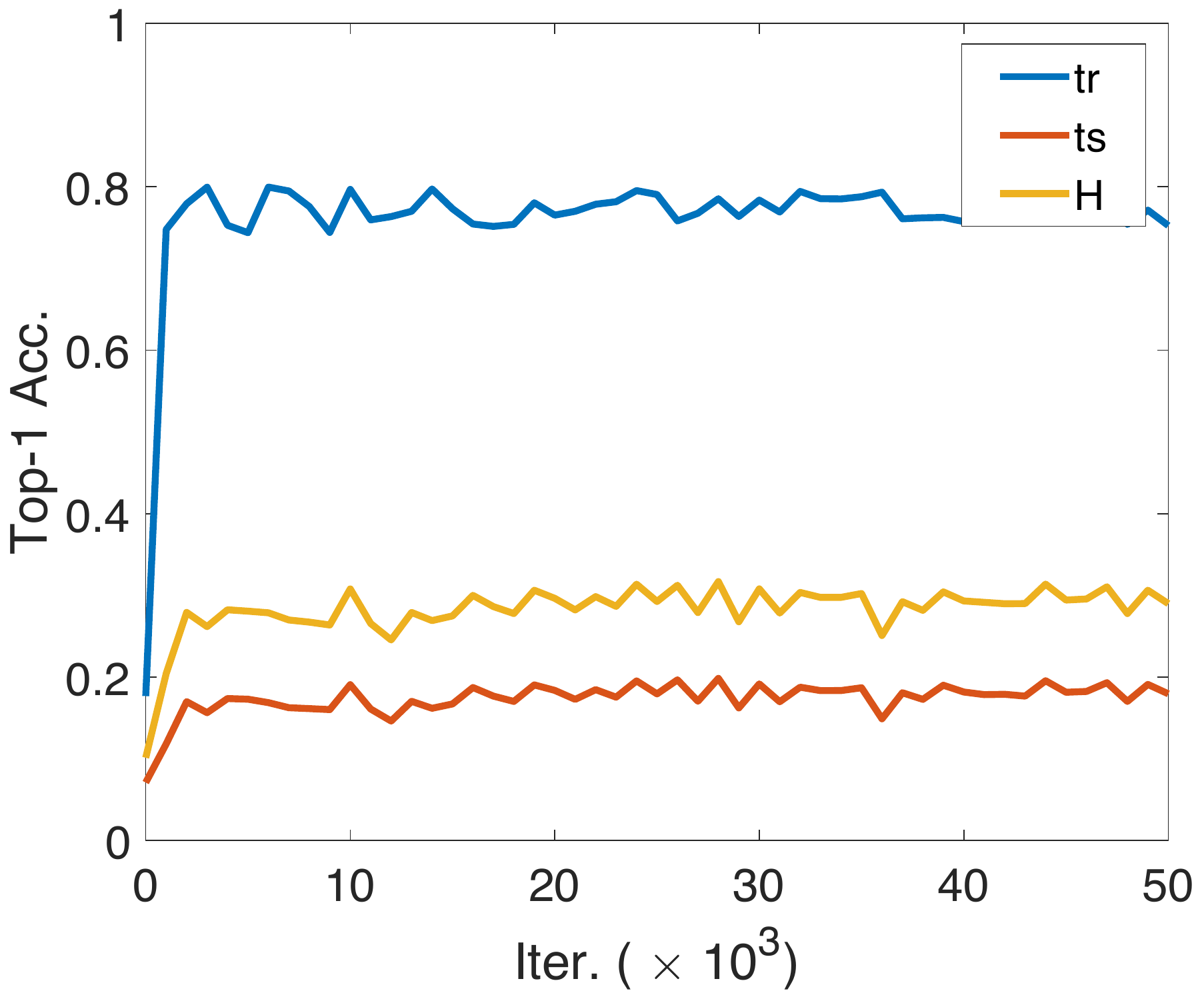}} %\vspace{-10pt}
	\vspace{-10pt}
	\caption{Testing accuracy against number of iterations on the tested five public datasets.}
	\label{fig:iteration}
	%\vspace{-10pt}
\end{figure*}

In fact, existing algorithms like DAP, CONSE and SYNC often achieve competitive results on seen classes, while poorly on unseen classes. These methods almost only consider seen classes like the one for traditional image classification, but the most important problem in ZSL is how to generalize seen classes to unseen classes, so these methods have limitations in this regard.

To illustrate the effectiveness of our method, we use the dataset AwA2 to visualize (by t-SNE \cite{maaten2008visualizing}) the embedding for unseen classes. As shown in Fig. \ref{fig:emb}, visual features of each unseen class are well separated from each other. We use black circles to indicate the embedding of unseen classes, and it is obvious that our approach embeds $80\%$ unseen classes into their own clusters. In contrast, the baseline DEM embed much fewer. DEM has difficulty in distinguishing the embedding between adjacent unseen classes, and as a result these embedding will gather together. By contrast, our algorithm can utilize neighbors to pull the embedding in different directions and separate them.

%We give a brief discussion on why our approach can separate these embedding effectively. For different classes, they have different neighbors, thus GNN can utilize these neighbors to pull the embedding in different directions and separate them. However, for the baseline DEM which does not utilize correlation between classes, the differences in attribute vectors are not sufficient to completely distinguish each other, thus the embedding of these unseen classes are easy to overlap.

%\vspace{-0.4cm}
%\smallskip
%\noindent {\bfseries{Ablation Analysis.}}
\subsubsection{Effectiveness of intra-class relationship.}
The aim of this study is to determine the effect of constructing complete-graph. On the left of Table \ref{tab:relation}, we find that without constructing complete-graph for each class, the results are lower than ones that do it. This indicates that construct intra-class relationship can improve result.
% Moreover, we observe that this improvement mainly depend on the seen classes, which demonstrate that our model can obtain better embedding for seen classes.  
% \usepackage{multirow}

\subsubsection{Effectiveness of inter-class relationship.}
We also evaluate our algorithm without constructing inter-class relationship. On the right of Table \ref{tab:relation}, we observe that constructing inter-class relationship outperforms the `none" by a large margin. In particular, for AwA2 dataset, the accuracy increases from 39.6\% to 52.7\%. These results show that our model can transfer knowledge from seen classes efficiently.

%\vspace{-0.3cm}
\begin{table}[tb!]
\centering
\caption{Effect of intra-class and inter-class relationship on five benchmarks, CG refer to the complete graph.}
\resizebox{0.8\textwidth}{!}{
\begin{tabular}{ccccc|cccc}
\hline
Dataset               & Intra-class    & $ts$    & $tr$    & $H$    & Inter-class & $ts$    & $tr$    & $H$    \\ \hline
\multirow{2}{*}{SUN}  & None           & 22.2 & 33.4 & 26.7 & None        & 21.5 & 32.5 & 25.9 \\
                      & CG & 22.3 & 36.5 & \textbf{27.7} & kNN         & 22.3 & 36.5 & \textbf{27.7} \\ \hline
\multirow{2}{*}{CUB}  & None           & 24.4 & 56.7 & 34.1 & None        & 23.0 & 56.6 & 32.7 \\
                      & CG & 25.2 & 56.9 & \textbf{34.9} & kNN         & 25.2 & 56.9 & \textbf{34.9} \\ \hline 
\multirow{2}{*}{AwA1} & None           & 34.9 & 82.6 & 49.1 & None        & 32.6 & 84.0 & 46.9 \\
                      & CG & 39.4 & 83.5 & \textbf{53.6} & kNN         & 39.4 & 83.5 & \textbf{53.6} \\ \hline
\multirow{2}{*}{AwA2} & None           & 36.1 & 85.5 & 50.7 & None        & 25.7 & 86.5 & 39.6 \\
                      & CG & 37.9 & 86.5 & \textbf{52.7} & kNN         & 37.9 & 86.5 & \textbf{52.7} \\ \hline
\multirow{2}{*}{aPY}  & None           & 14.3 & 56.7 & 22.8 & None        & 13.1 & 76.3 & 22.4 \\
                      & CG & 18.3 & 79.0 & \textbf{29.7} & kNN         & 18.3 & 79.0 & \textbf{29.7} \\ \hline
\end{tabular}}
\label{tab:relation}
\end{table}

\subsubsection{Effectiveness of Wasserstein metric.}
We perform ablation study to the efficiency of different node selection methods in our approach. Besides the Wasserstein barycenter, we can also select representative nodes by Euclidean barycenter (averaging directly) and random number. As shown in Fig. \ref{fig:acc}, the Wasserstein barycenter outperforms the others. It is obvious that selecting representative nodes randomly usually causes a significant drop in accuracy, for the reason that the resulting representative node is more likely to fall into the overlap, thus it is prone to classify images into wrong class by using kNN during testing. Moreover, by visualizing these representative nodes selected by different techniques in Fig. \ref{fig:animals}, we can find that intuitively the nodes selected by Wasserstein distance have more distinctive features.

%In addition, the former usually reflects the obvious features of an object, while the features of latter are sometimes inconspicuous. For example, the representative node selected by Euclidean distance for gorilla is a child whose physical characteristics have not yet fully developed, for killer whale, there are only fins out of the water in the representative image, thus these representative node can not represent for the whole class well. However, 
%In addition, it is interesting that the images selected by Wasserstein barycenter can highlight the characteristics of the class. such as for antelope and moose, the horns and antlers are the most recognizable features, so these features occupy a large proportion in the selected images.

\noindent {\bfseries{Convergence and Scalability.}}
The accuracy on the test set against the number of iterations for the five benchmark datasets is shown in Fig. \ref{fig:iteration}. It is also noteworthy that our approach can quickly converge to the final results on all datasets. In addition, in order to address the scalability issues, we uniformly sample a fixed-set of neighbors $S$, instead of using full neighborhood sets during training. In addition, because the depth of our algorithm is 2, thus the per-batch time complexity is fixed at $O(S^2)$. In experiments, we set $S$ equals to 50. 

Moreover, due to the large amount of data, we choose Sinkhorn distance \cite{cuturi2013sinkhorn} to compute an approximate Wasserstein barycenter, which also possess the scalability problem.
Specifically, the penalty weight for each instance is set to be same, the cost matrix $C$ is set to be $C_{ij} = \left\| i-j \right\|_2^2$ and the parameter $\epsilon$ in Sinkhorn is set to be $1e^{-5}$, while guarantees estimated good enough results.

%\vspace{-0.3cm}
\section{Conclusion}\label{sec:con}
We have presented a heterogeneous graph-based model for GZSL. Unlike most existing algorithms, our method is agnostic to test data during training, which makes our approach can adapt to the dynamic scenarios effectively. We construct a meaningful structured graph to represent the relation among the data.
Instead of Euclidean space, our experiments show that constructing graph in Wasserstein space can achieve better results. Further the graph neural network can be employed to train our model on the well-constructed graph. Extensive experimental results show that more explainable embedding can be obtained for unseen classes which leads to higher top-$1$ accuracy compared to peer methods. %In the near future, we will extend the experiment to the ImageNet dataset and then consider the end-to-end implementation.

%\section{Acknowledgments}
%\clearpage
\small
\bibliographystyle{plain}
%\balance
\bibliography{GZSL_v1}

\end{document}